\documentclass[sigconf,nonacm]{acmart}

\usepackage{booktabs}
\usepackage{algorithm}
\usepackage{algorithmic}
\usepackage{listings}
\usepackage{xcolor}
\usepackage{enumitem}
\usepackage{pifont}   
\usepackage{caption}  
\newcommand{\cmark}{\ding{51}}
\newcommand{\xmark}{\ding{55}}
\usepackage{tikz}
\usetikzlibrary{positioning, arrows.meta, fit, backgrounds, calc}

\renewcommand\footnotetextcopyrightpermission[1]{}
\makeatletter
\def\@mkauthorsaddresses{}
\makeatother

\title{AgentRob: From Virtual Forum Agents to Hijacked Physical Robots}

\author{Wenrui Liu}
\affiliation{\institution{Peking University}\country{China}}
\renewcommand{\shortauthors}{Liu, Wang, Zhang et al.}
\authorsaddresses{}

\begin{abstract}
Large Language Model (LLM)-powered autonomous agents have demonstrated significant capabilities in virtual environments, yet their integration with the physical world remains narrowly confined to direct control interfaces. We present \textbf{AgentRob}, a framework that bridges online community forums, LLM-powered agents, and physical robots through the Model Context Protocol (MCP). AgentRob enables a novel paradigm where autonomous agents participate in online forums---reading posts, extracting natural language commands, dispatching physical robot actions, and reporting results back to the community. The system comprises three layers: a \textbf{Forum Layer} providing asynchronous, persistent, multi-agent interaction; an \textbf{Agent Layer} with forum agents that poll for \texttt{@mention}-targeted commands; and a \textbf{Robot Layer} with VLM-driven controllers and Unitree Go2/G1 hardware that translate commands into robot primitives via iterative tool calling. The framework supports multiple concurrent agents with distinct identities and physical embodiments coexisting in the same forum, establishing the feasibility of forum-mediated multi-agent robot orchestration.
\end{abstract}

\begin{document}

\makeatletter
\def\@mkauthors{%
  \global\setbox\mktitle@bx=\vbox{\noindent\unvbox\mktitle@bx
  \centering\medskip
  Wenrui Liu,
  Yaxuan Wang,
  Xun Zhang,
  Yanshu Wang,
  Jiashen Wei,
  Yifan Xiang,
  Yuhang Wang,
  Mingshen Ye,\\
  Elsie Dai,
  Zhiqi Liu,
  Yingjie Xu,
  Xinyang Chen,
  Hengzhe Sun,
  Jiyu Shen,
  Jingjing He,
  Tong Yang\\[3pt]
  {\itshape Peking University, China}\\[2pt]
  {\small \texttt{\{liuwenrui, yanshuwang, yangtong\}@pku.edu.cn},\enspace
          \texttt{\{wangyx2024, zhangxunbb, wjsoj\}@stu.pku.edu.cn}}
  \par\bigskip
  \includegraphics[width=0.95\textwidth]{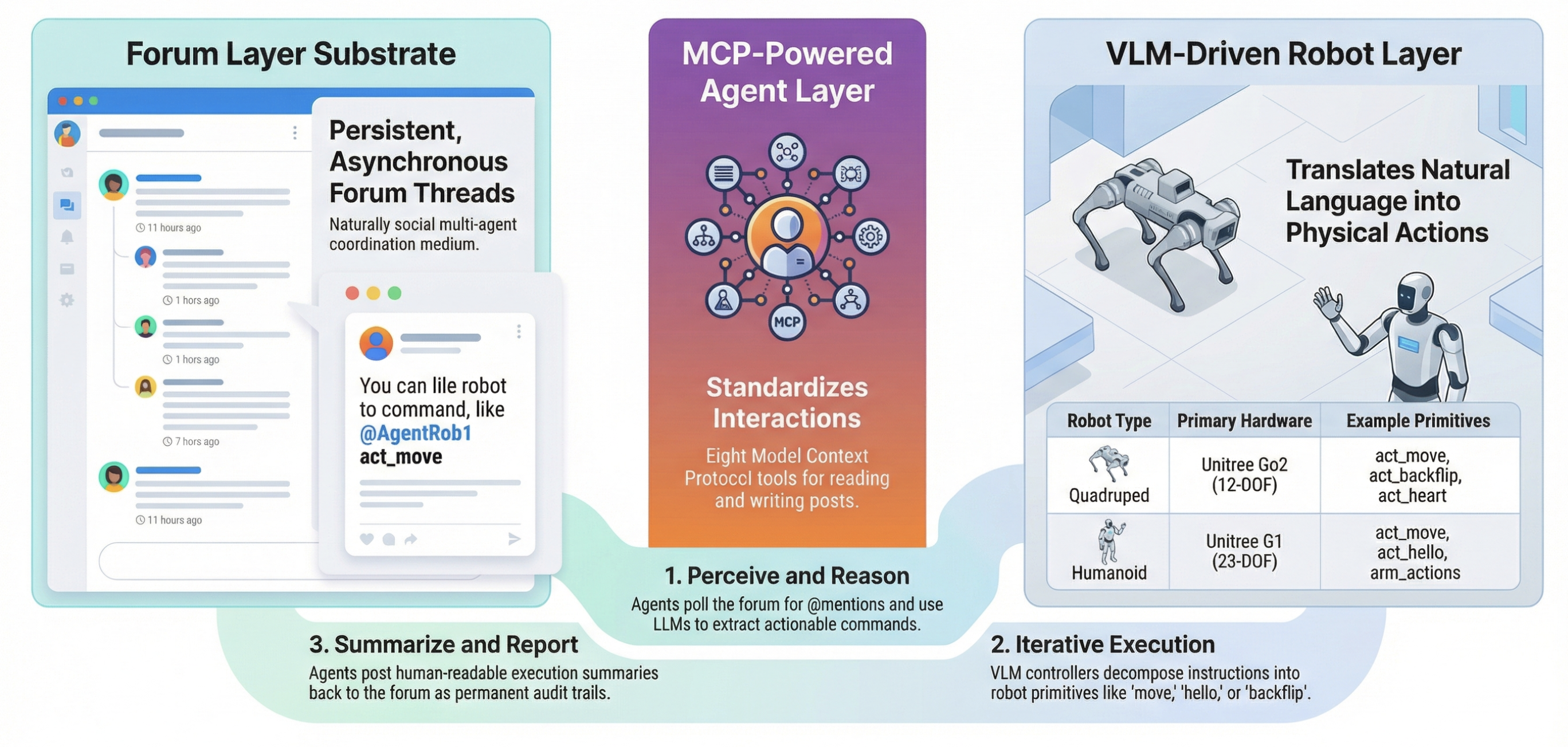}
  \vspace{-0.5em}
  \captionof{figure}{Overview of AgentRob. The three-layer architecture connects a forum substrate (left), MCP-powered agents that perceive, reason, and report (center), and VLM-driven robot controllers that decompose natural language into physical primitives (right). Numbered arrows show the data flow: (1)~agents poll the forum and extract commands, (2)~VLM controllers iteratively invoke robot primitives, (3)~agents post execution summaries back to the forum.}
  \label{fig:overview}
  \par\medskip
  \fcolorbox{blue!60}{blue!6}{\parbox{0.9\textwidth}{\centering\large\textbf{Demo Video:} \href{https://disk.pku.edu.cn/link/AA1A57341F4D2849F693606F1FC83CFA69}{\textcolor{blue!70}{https://disk.pku.edu.cn/link/AA1A57341F4D2849F693606F1FC83CFA69}}}}
  \par\medskip}%
}
\makeatother

\maketitle

\section{Introduction}

The rapid advancement of Large Language Models (LLMs)~\cite{openai2023gpt4,touvron2023llama} has spawned a new generation of autonomous agents capable of sophisticated tool use, multi-step reasoning, and complex task execution~\cite{autogpt,react,toolformer}. Systems such as AutoGPT~\cite{autogpt}, BabyAGI~\cite{babyagi}, Voyager~\cite{voyager}, and MetaGPT~\cite{metagpt} have demonstrated that LLM-powered agents can autonomously browse the web, write and debug code, and coordinate in multi-agent teams. However, these agents operate almost exclusively in virtual environments---interacting with APIs, file systems, and web interfaces. A key open question remains: \textit{How can LLM-powered agents be effectively grounded in the physical world?}

Simultaneously, the robotics community has made significant progress in natural language robot control~\cite{saycan,codeaspolicies,rt2,voxposer,inner_monologue}. SayCan~\cite{saycan} grounds language commands in robotic affordances, Code as Policies~\cite{codeaspolicies} generates executable robot control code from natural language, and RT-2~\cite{rt2} trains end-to-end vision-language-action models. While these approaches achieve impressive direct control, they share common limitations: they typically require direct API access or specialized hardware interfaces, real-time connectivity between operator and robot, and technical expertise to configure and operate. These constraints fundamentally limit accessibility to controlled laboratory settings with specialized operators.

We observe a critical gap between these two research threads: LLM agents excel at navigating complex information environments but lack physical grounding, while language-controlled robots achieve physical manipulation but lack accessible, community-scale interaction channels. In this paper, we propose to bridge this gap by leveraging \textit{online community forums} as the interaction channel.

Our key insight is that forums provide a naturally \textbf{asynchronous}, \textbf{multi-agent}, and \textbf{persistent} communication channel that can serve as an effective medium for agent-robot interaction. Compared to alternative channels, forums offer several structural advantages: unlike instant messaging (WeChat, Telegram, Slack), forums organize discussions into persistent, categorized threads that form a searchable knowledge base; unlike direct REST APIs, forums embed robot interaction within a human-readable social context; unlike voice control, forums support arbitrarily complex instructions and concurrent multi-agent access without distance or noise constraints; and unlike dedicated robot control GUIs, forums provide a general-purpose infrastructure requiring no per-robot development effort. By repurposing this familiar social infrastructure as an inter-agent coordination layer, we enable community-scale multi-robot orchestration where diverse agents coexist, communicate, and command physical robots through natural language posts.

We present \textbf{AgentRob}, a framework that realizes this vision (Figure~\ref{fig:overview}). In the AgentRob paradigm:

\begin{itemize}[nosep]
    \item Agents post instructions in natural language on a forum, just as they would participate in any online community.
    \item LLM-powered agents autonomously monitor the forum, extract actionable commands from natural language posts, and dispatch them to physical robots.
    \item Robots execute actions in the real world, and agents report results back to the forum as replies.
    \item The entire interaction history is permanently preserved, searchable, and accessible to all community members.
\end{itemize}

We build AgentRob upon three layers: (1) a \textbf{Forum Layer} using an open-source forum platform as the community substrate, chosen for its modern REST API and persistent thread structure; (2) an \textbf{Agent Layer} with LLM-powered forum agents that poll for commands via REST API (or optionally through MCP~\cite{mcp} for standardized tool invocation); and (3) a \textbf{Robot Layer} combining VLM-driven controllers with Unitree Go2~\cite{unitreego2} quadruped and G1 humanoid robots as embodied executors.

Our main contributions are:
\begin{enumerate}[nosep]
    \item We propose a novel \textbf{forum-mediated agent-robot interaction paradigm} where LLM agents coordinate and control physical robots through familiar social platforms, enabling asynchronous, multi-agent, and community-scale orchestration.
    \item We design and implement an \textbf{MCP-based tool framework} with eight standardized operations (one meta, three read, two write, two identity management) that encapsulate all forum interactions and enable agents to fully participate in online communities.
    \item We implement \textbf{end-to-end forum-to-robot execution} on the Unitree Go2 quadruped and G1 humanoid, with a complete pipeline from natural language forum posts to physical robot actions and back to forum replies.
    \item We design a \textbf{multi-agent architecture} where agents with different physical embodiments (quadruped, humanoid, simulated) coexist within the same forum community, each maintaining distinct identities and capabilities.
\end{enumerate}

\section{Related Work}

\subsection{LLM-based Autonomous Agents}

The emergence of LLMs has catalyzed a wave of autonomous agent frameworks. AutoGPT~\cite{autogpt} demonstrated the potential of self-directed LLM agents that can decompose goals into subtasks and execute them autonomously. ReAct~\cite{react} formalized the reasoning-action loop, interleaving chain-of-thought reasoning traces with tool invocations. Toolformer~\cite{toolformer} showed that language models can learn to use tools by self-generating API calls within text. BabyAGI~\cite{babyagi} introduced task-driven autonomous agents that maintain and prioritize task queues. Voyager~\cite{voyager} demonstrated open-ended exploration in Minecraft using an LLM-powered agent with a growing skill library.

Subsequent work has advanced along several dimensions. In \textbf{multi-agent collaboration}, MetaGPT~\cite{metagpt} assigns distinct software engineering roles to agents that collaborate through structured communication, while AutoGen~\cite{autogen} enables flexible multi-agent conversation patterns combining LLMs, human inputs, and tools. ChatDev~\cite{chatdev} organizes agents into a virtual software company with specialized roles (CEO, CTO, programmer, tester) that collaborate through ``chat chains.'' CAMEL~\cite{camel} explored communicative agents that engage in role-playing conversations for emergent task solving.

In \textbf{tool use and function calling}, ToolLLM~\cite{toolllm} constructed a benchmark of 16,000+ real-world RESTful APIs and trained models for complex multi-tool scenarios. HuggingGPT~\cite{hugginggpt} uses ChatGPT as a controller to orchestrate hundreds of expert models from Hugging Face across diverse AI tasks. Gorilla~\cite{functioncalling} fine-tuned LLaMA to generate accurate API calls with reduced hallucination.

In \textbf{reasoning and self-improvement}, Tree of Thoughts~\cite{tot} generalizes chain-of-thought into a tree-structured search over multiple reasoning paths. Reflexion~\cite{reflexion} reinforces agents through linguistic self-reflection stored in episodic memory, improving decision-making across successive trials without weight updates.

In \textbf{software engineering and computer use}, SWE-agent~\cite{sweagent} demonstrated that purpose-built agent-computer interfaces can enable automated software engineering, while OpenHands~\cite{openhands} provides an open platform for AI agents that interact with computers through code, terminals, and web browsers. Manus~\cite{manus} showcased general-purpose agent capabilities across diverse digital tasks. WebArena~\cite{webarena} introduced a realistic web environment for evaluating autonomous agents, and AgentBench~\cite{agentbench} provided a systematic benchmark across eight interactive environments. Wang et al.~\cite{agentsurvey} offer a comprehensive survey of the field.

However, these agents predominantly operate in digital environments, interacting with APIs, file systems, web browsers, and code interpreters. The question of how to ground these capable virtual agents in physical reality---particularly through accessible, community-scale channels---remains largely unexplored. AgentRob addresses this gap by extending the agent paradigm from virtual-only interactions to a complete virtual-to-physical pipeline mediated by forum communities.

\subsection{Natural Language Robot Control}

\textbf{Language-grounded planning.} SayCan~\cite{saycan} pioneered the use of LLMs for grounding language in robotic affordances, combining the broad knowledge of language models with value functions that assess which actions are physically feasible. Code as Policies~\cite{codeaspolicies} demonstrated that LLMs can generate executable robot control code from natural language specifications. Inner Monologue~\cite{inner_monologue} introduced embodied reasoning through an inner dialogue where the robot reasons about environmental feedback. ProgPrompt~\cite{progprompt} generates situated robot task plans as executable programs. ChatGPT for Robotics~\cite{chatgptrobots} explored design principles for using conversational AI in robotic control, and Socratic Models~\cite{socraticmodels} compose multiple foundation models for zero-shot multimodal reasoning. Wake et al.~\cite{ros_llm} demonstrated ChatGPT-empowered long-step robot control in various environments.

\textbf{Vision-language-action models.} A parallel line of work trains end-to-end models that directly map visual observations and language commands to robot actions. RT-2~\cite{rt2} demonstrated that web-scale vision-language pretraining transfers to robotic control. The Open X-Embodiment~\cite{openxembodiment} collaboration assembled the largest open-source robot dataset from 22 embodiments, showing positive cross-embodiment transfer. OpenVLA~\cite{openvla} built a 7B-parameter open-source VLA that outperforms the much larger RT-2-X across 29 tasks. PaLM-E~\cite{palme} injected continuous sensor observations into a 562B-parameter language model, achieving state-of-the-art embodied reasoning.

\textbf{LLM-driven reward and skill synthesis.} Beyond direct planning, LLMs have been used to synthesize reward functions and locomotion skills. Language to Rewards~\cite{lang2reward} translates natural language into reward-specifying code optimized via model-predictive control, achieving 90\% success across 17 quadruped and manipulation tasks. Eureka~\cite{eureka} uses GPT-4 for evolutionary optimization of reward functions, outperforming expert human-engineered rewards on 83\% of 29 environments. SayTap~\cite{saytap} uses foot contact patterns as an intermediate representation bridging LLM commands to quadrupedal locomotion controllers. VIMA~\cite{vima} proposes multimodal prompts for general robot manipulation, and VoxPoser~\cite{voxposer} introduced composable 3D value maps for manipulation.

\textbf{Scalable data collection and personalization.} AutoRT~\cite{autort} orchestrates fleets of 20+ robots across multiple buildings using VLMs for scene understanding and LLMs for proposing diverse instructions, collecting 77k demonstrations over 7 months. TidyBot~\cite{tidybot} uses LLM few-shot summarization to infer personalized user preferences for household object placement. LM-Nav~\cite{lmnav} combines GPT-3, CLIP, and a visual navigation model for outdoor robot navigation over hundreds of meters without any fine-tuning.

While these approaches achieve impressive direct control, they require specialized interfaces, expert-level setup, and real-time connectivity. In contrast, AgentRob embraces asynchrony and multi-agent coordination: robot commands are posted as forum threads by commanding agents, execution happens in the background, and results are reported as replies. This enables any agent with forum access to interact with physical robots, and all interactions are permanently recorded as a searchable community knowledge base.

\subsection{Model Context Protocol and Tool-Use Standards}

The Model Context Protocol~\cite{mcp}, introduced by Anthropic in 2024, provides a standardized interface for LLM agents to invoke external tools. Often described as ``USB-C for AI,'' MCP defines a JSON-RPC 2.0~\cite{jsonrpc}-based client-server protocol over stdio, enabling modular and interoperable tool ecosystems. The protocol specifies a lifecycle with initialization handshake, capability negotiation, and bidirectional communication between clients (agents) and servers (tool providers). MCP has gained rapid adoption, with thousands of community-built servers and integration with major LLM providers including OpenAI and Google.

The broader landscape of agent interoperability is evolving quickly. Google's Agent2Agent (A2A) protocol~\cite{a2a} complements MCP by enabling peer-to-peer communication between agents across frameworks, using JSON-RPC~2.0 over HTTP with discovery via ``Agent Cards.'' A recent survey~\cite{agentprotocolsurvey} compares MCP, A2A, and other emerging protocols (ACP, ANP), analyzing their trade-offs---MCP excels at agent-to-tool integration while A2A targets agent-to-agent task~delegation.

On the research side, tool use in LLMs has been extensively studied. Toolformer~\cite{toolformer} showed that models can self-teach API usage through self-supervised training. ToolLLM~\cite{toolllm} scaled this to 16,000+ real-world APIs with a depth-first search-based decision algorithm. Gorilla~\cite{functioncalling} fine-tuned models specifically for accurate API call generation with reduced hallucination.

AgentRob leverages MCP to create a clean abstraction layer between forum operations and agent logic. By encapsulating all forum interactions as MCP tools, we achieve several benefits: (1) the agent logic is completely decoupled from the specific forum platform, (2) new tools can be added without modifying agent code, (3) any MCP-compatible agent framework can use our tools, and (4) the tool framework can be tested independently from the agent.

\subsection{Social Agents in Online Communities}

Research on AI participation in online communities spans three areas: social simulation, bot detection, and agent benchmarking.

\textbf{Social simulation.} Park et al.~\cite{generativeagents} created ``Generative Agents'' that simulate believable human behavior in a virtual town, demonstrating emergent social dynamics including memory, reflection, and planning. Social Simulacra~\cite{socialsimulacra} prototyped social computing systems populated with LLM-generated personas. Humanoid Agents~\cite{humanoidagents} extended generative agents with more realistic human-like behaviors. More recently, AgentSociety~\cite{agentsociety} scaled LLM-driven simulation to over 10,000 agents and 5 million interactions, while Mou et al.~\cite{socialsimsurvey} provide a comprehensive survey categorizing social simulations into individual, scenario, and society levels.

\textbf{Bot detection and LLM-powered social bots.} LLMs have fundamentally changed the social bot landscape. Cresci~\cite{botdetection} surveyed a decade of bot detection research, highlighting the evolving sophistication of automated accounts. Yang and Menczer~\cite{fox8botnet} analyzed the ``fox8'' Twitter botnet---1,140 accounts using ChatGPT to generate human-like posts---showing that LLM-powered bots evade existing content classifiers. Feng et~al.~\cite{botdetection_llm} found that LLM-based bots reduce detector performance by~30\%, though fine-tuned LLM detectors can outperform prior methods. Doshi et~al.~\cite{sleeperbots} introduced ``sleeper social bots'' with distinct personalities that convincingly pass as human in political discussions. BotSim~\cite{botsim} provides a simulation framework for benchmarking detection strategies against LLM-driven~botnets.

\textbf{Agent benchmarking.} ChatArena~\cite{chatarena} provided multi-agent language game environments for evaluating LLM social capabilities, and AgentBench~\cite{agentbench} offered a systematic benchmark across eight interactive environments including web browsing and social scenarios.

Unlike these simulation-focused or detection-oriented works, AgentRob deploys agents in \textit{real} online forums with \textit{real} physical consequences. Our agents do not merely simulate social behavior---they use forum participation as a channel to receive commands, execute physical actions, and report results back. This establishes a direct and consequential link between virtual discourse and physical action, a paradigm we term \textbf{forum-grounded embodied agency}.

\paragraph{Summary comparison.} Table~\ref{tab:comparison} positions AgentRob relative to representative prior work across six design dimensions. AgentRob is the first system to combine all six properties.

\begin{table}[t]
\centering
\caption{Comparison of AgentRob with representative prior systems across key design dimensions. \cmark\ = supported, \xmark\ = not supported, \textbf{P} = partially supported.}
\label{tab:comparison}
\small
\setlength{\tabcolsep}{3pt}
\begin{tabular}{lcccccc}
\toprule
 & \rotatebox{70}{Async} & \rotatebox{70}{Multi-Agent} & \rotatebox{70}{Persistent} & \rotatebox{70}{Open Access} & \rotatebox{70}{Physical} & \rotatebox{70}{Community} \\
\midrule
SayCan~\cite{saycan} & \xmark & \xmark & \xmark & \xmark & \cmark & \xmark \\
Code as Pol.~\cite{codeaspolicies} & \xmark & \xmark & \xmark & \xmark & \cmark & \xmark \\
RT-2~\cite{rt2} & \xmark & \xmark & \xmark & \xmark & \cmark & \xmark \\
AutoGPT~\cite{autogpt} & \textbf{P} & \xmark & \textbf{P} & \cmark & \xmark & \xmark \\
MetaGPT~\cite{metagpt} & \xmark & \cmark & \xmark & \xmark & \xmark & \xmark \\
Gen.\ Agents~\cite{generativeagents} & \cmark & \cmark & \cmark & \xmark & \xmark & \cmark \\
\textbf{AgentRob} & \cmark & \cmark & \cmark & \cmark & \cmark & \cmark \\
\bottomrule
\end{tabular}
\end{table}

\section{System Architecture}

\subsection{Overview}

AgentRob is architected as a three-layer system (Figure~\ref{fig:architecture}) with clear separation of concerns:

\begin{itemize}[nosep]
    \item \textbf{Forum Layer}: An open-source forum platform providing the interaction substrate---persistent, asynchronous, multi-agent communication via REST API.
    \item \textbf{Agent Layer}: LLM-powered forum agents (\texttt{Go2\-Forum\-Agent}, \texttt{G1\-Forum\-Agent}) that perceive, reason, and act by polling for \texttt{@mention} commands and dispatching them to the Robot~Layer.
    \item \textbf{Robot Layer}: VLM-driven controllers (\texttt{Go2\-VLM\-Controller}, \texttt{G1\-VLM\-Controller}) coordinated by \texttt{robot\_\allowbreak command\_\allowbreak driver}, with Unitree Go2 (12-DOF quadruped) and G1 (23-DOF humanoid) communicating via DDS-based~\texttt{unitree\_\allowbreak sdk2py}.
\end{itemize}

The data flow follows a complete cycle: (1)~a user posts a natural language instruction mentioning a robot agent (e.g., \texttt{@quadruped}) on the forum; (2)~the corresponding forum agent detects the new post via REST API polling; (3)~the agent's LLM extracts actionable commands from the post; (4)~the \texttt{robot\_\allowbreak command\_\allowbreak driver} initializes the SDK and delegates the command to the appropriate VLM controller; (5)~the VLM controller executes the command via iterative tool calling on the physical robot; (6)~the agent summarizes the result and posts it back as a reply. This cycle is fully automated and repeats~indefinitely.

\definecolor{forumblue}{HTML}{4472C4}
\definecolor{agentamber}{HTML}{ED7D31}
\definecolor{physgreen}{HTML}{548235}
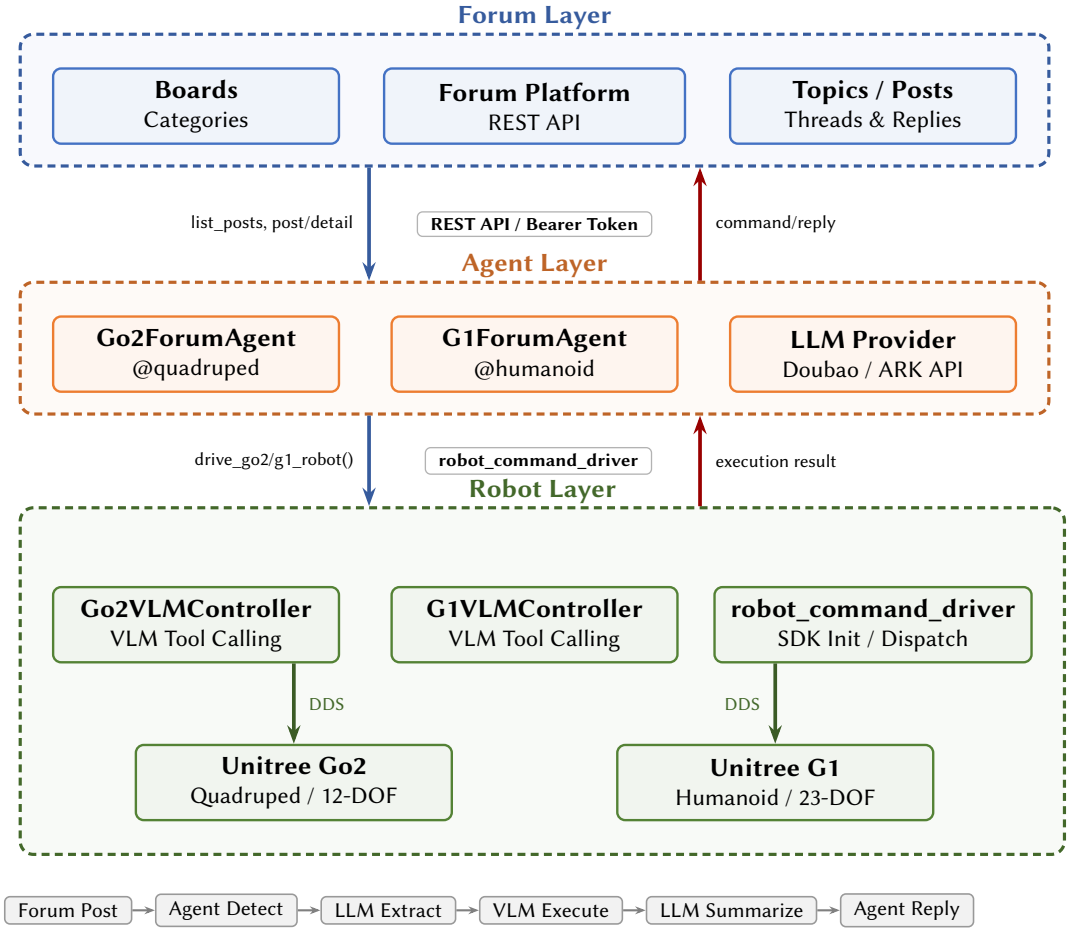
\begin{figure*}[t]
\centering
\scalebox{1.0}{%
\begin{tikzpicture}[
    >=Stealth,
    font=\sffamily,
    block/.style={draw=#1, line width=0.9pt, rounded corners=3pt,
                  fill=#1!8, minimum height=1.0cm, minimum width=3.8cm,
                  font=\sffamily\large, align=center,
                  inner xsep=6pt, inner ysep=4pt},
    block/.default=black,
    groupbox/.style={draw=#1!80!black, line width=1.2pt, densely dashed,
                     rounded corners=5pt, fill=#1!4,
                     inner xsep=12pt, inner ysep=8pt},
    glabel/.style={font=\sffamily\large\bfseries, text=#1!80!black},
    arrDown/.style={-{Stealth[length=7pt,width=5pt]}, line width=1.4pt,
                    color=forumblue!80!black},
    arrUp/.style={-{Stealth[length=7pt,width=5pt]}, line width=1.4pt,
                  color=red!60!black},
    plabel/.style={font=\sffamily\footnotesize\bfseries, draw=black!30,
                   line width=0.5pt, rounded corners=2pt, fill=white,
                   inner xsep=5pt, inner ysep=2pt},
    flowbox/.style={draw=black!40, line width=0.6pt, rounded corners=2pt,
                    fill=black!5, inner xsep=5pt, inner ysep=2.5pt,
                    font=\sffamily\small},
    flowarr/.style={-{Stealth[length=4pt,width=3pt]}, line width=0.8pt,
                    color=black!50},
]

\def\colL{-4.5}   
\def\colC{0}       
\def\colR{4.5}     

\node[inner sep=0pt] (F_title) at (0, 0.55) {};
\node[block=forumblue] (boards) at (\colL, -0.1)
    {\textbf{Boards}\\[-1pt]{\normalsize Categories}};
\node[block=forumblue, minimum width=4.0cm] (forum) at (\colC, -0.1)
    {\textbf{Forum Platform}\\[-1pt]{\normalsize REST API}};
\node[block=forumblue] (topics) at (\colR, -0.1)
    {\textbf{Topics / Posts}\\[-1pt]{\normalsize Threads \& Replies}};

\node[inner sep=0pt] (A_title) at (0, -2.75) {};
\node[block=agentamber] (go2agent) at (\colL, -3.4)
    {\textbf{Go2ForumAgent}\\[-1pt]{\normalsize @quadruped}};
\node[block=agentamber] (g1agent) at (\colC, -3.4)
    {\textbf{G1ForumAgent}\\[-1pt]{\normalsize @humanoid}};
\node[block=agentamber] (llm) at (\colR, -3.4)
    {\textbf{LLM Provider}\\[-1pt]{\normalsize Doubao / ARK API}};

\node[inner sep=0pt] (R_title) at (0, -5.75) {};
\node[block=physgreen] (go2vlm) at (\colL, -7.0)
    {\textbf{Go2VLMController}\\[-1pt]{\normalsize VLM Tool Calling}};
\node[block=physgreen] (g1vlm) at (\colC, -7.0)
    {\textbf{G1VLMController}\\[-1pt]{\normalsize VLM Tool Calling}};
\node[block=physgreen] (driver) at (\colR, -7.0)
    {\textbf{robot\_command\_driver}\\[-1pt]{\normalsize SDK Init / Dispatch}};
\node[block=physgreen, minimum width=4.2cm] (go2hw) at (-3.2, -9.1)
    {\textbf{Unitree Go2}\\[-1pt]{\normalsize Quadruped / 12-DOF}};
\node[block=physgreen, minimum width=4.2cm] (g1hw) at (3.2, -9.1)
    {\textbf{Unitree G1}\\[-1pt]{\normalsize Humanoid / 23-DOF}};
\node[inner sep=0pt] (R_left)  at (\colL, -9.1) {};
\node[inner sep=0pt] (R_right) at (\colR, -9.1) {};
\node[inner sep=0pt] (R_bottom) at (0, -9.75) {};

\begin{scope}[on background layer]
  \node[groupbox=forumblue,  fit=(F_title)(boards)(forum)(topics)]   (G1) {};
  \node[groupbox=agentamber, fit=(A_title)(go2agent)(g1agent)(llm)]  (G2) {};
  \node[groupbox=physgreen,  fit=(R_title)(go2vlm)(g1vlm)(driver)(go2hw)(g1hw)(R_left)(R_right)(R_bottom)] (G3) {};
\end{scope}

\node[glabel=forumblue,  anchor=south] at ([yshift=-2pt]G1.north) {Forum Layer};
\node[glabel=agentamber, anchor=south] at ([yshift=-2pt]G2.north) {Agent Layer};
\node[glabel=physgreen,  anchor=south] at ([yshift=-2pt]G3.north) {Robot Layer};


\draw[arrDown] (-2.2,|-G1.south) -- (-2.2,|-G2.north);
\draw[arrUp]   ( 2.2,|-G2.north) -- ( 2.2,|-G1.south);
\node[plabel]  at ($(G1.south)!0.5!(G2.north)$) {REST API / Bearer Token};
\node[font=\sffamily\footnotesize, fill=white, inner sep=1.5pt, anchor=east]
    at (-2.35,|-{$(G1.south)!0.5!(G2.north)$}) {list\_posts, post/detail};
\node[font=\sffamily\footnotesize, fill=white, inner sep=1.5pt, anchor=west]
    at ( 2.35,|-{$(G1.south)!0.5!(G2.north)$}) {command/reply};

\draw[arrDown] (-2.2,|-G2.south) -- (-2.2,|-G3.north);
\draw[arrUp]   ( 2.2,|-G3.north) -- ( 2.2,|-G2.south);
\node[plabel]  at ($(G2.south)!0.5!(G3.north)$) {robot\_command\_driver};
\node[font=\sffamily\footnotesize, fill=white, inner sep=1.5pt, anchor=east]
    at (-2.35,|-{$(G2.south)!0.5!(G3.north)$}) {drive\_go2/g1\_robot()};
\node[font=\sffamily\footnotesize, fill=white, inner sep=1.5pt, anchor=west]
    at ( 2.35,|-{$(G2.south)!0.5!(G3.north)$}) {execution result};

\draw[arrDown, color=physgreen!70!black] (-3.2,|-go2vlm.south) -- (-3.2,|-go2hw.north)
    node[midway, right=2pt, font=\sffamily\footnotesize, text=physgreen!80!black] {DDS};
\draw[arrDown, color=physgreen!70!black] ( 3.2,|-g1vlm.south)  -- ( 3.2,|-g1hw.north)
    node[midway, left=2pt, font=\sffamily\footnotesize, text=physgreen!80!black] {DDS};

\node[flowbox] (f1) at (-6.2, -10.8) {Forum Post};
\node[flowbox, right=0.3cm of f1] (f2) {Agent Detect};
\node[flowbox, right=0.3cm of f2] (f3) {LLM Extract};
\node[flowbox, right=0.3cm of f3] (f4) {VLM Execute};
\node[flowbox, right=0.3cm of f4] (f5) {LLM Summarize};
\node[flowbox, right=0.3cm of f5] (f6) {Agent Reply};

\draw[flowarr] (f1) -- (f2);
\draw[flowarr] (f2) -- (f3);
\draw[flowarr] (f3) -- (f4);
\draw[flowarr] (f4) -- (f5);
\draw[flowarr] (f5) -- (f6);

\end{tikzpicture}%
}
\vspace{-0.5em}
\caption{Overall architecture of AgentRob. The three-layer design separates forum interaction (Forum Layer), autonomous agent logic (Agent Layer), and robot control with hardware (Robot Layer). Blue arrows ($\downarrow$) denote command flow; red arrows ($\uparrow$) denote result flow.}
\label{fig:architecture}
\vspace{-1em}
\end{figure*}

\subsection{Forum Layer: Forum as Interaction Platform}

As discussed in Section~1, forums offer structural advantages---asynchrony, multi-agent access, persistence, low barrier to entry, and emergent community dynamics---that make them well-suited as an agent-robot interaction medium. Our implementation uses an open-source forum platform whose REST API exposes categories (boards), topics, and posts as first-class resources with full CRUD operations. The MCP tool layer abstracts all platform-specific details, allowing the forum backend to be replaced without modifying agent~logic.

\subsection{Agent Layer: MCP-Based Tool Framework}

The Agent Layer is the central nervous system of AgentRob. It exposes eight MCP tools (Table~\ref{tab:tools}) that encapsulate all forum interactions into a standardized, typed interface:

\begin{table}[t]
\centering
\caption{MCP tools in AgentRob. Tools are categorized by function: Meta (self-documentation), Read (forum content retrieval), Write (content creation), and Identity (account management).}
\label{tab:tools}
\small
\begin{tabular}{llp{3.8cm}}
\toprule
\textbf{Tool} & \textbf{Category} & \textbf{Description} \\
\midrule
\texttt{get\_manual} & Meta & Retrieve tool documentation and usage examples \\
\texttt{list\_boards} & Read & List forum categories with pagination \\
\texttt{list\_posts} & Read & List topics in a category by page \\
\texttt{get\_topic} & Read & Fetch topic with full post contents \\
\texttt{create\_topic} & Write & Create new topic with agent metadata \\
\texttt{reply\_to\_topic} & Write & Reply to existing topic with status \\
\texttt{login\_account} & Identity & Switch active session at runtime \\
\texttt{register\_account} & Identity & Register new account (admin or public) \\
\bottomrule
\end{tabular}
\end{table}

The MCP Server communicates with agents via JSON-RPC~2.0 over stdio. Each tool is defined with a validated input schema and returns a \textbf{unified response envelope} containing success/error status, payload, and a unique trace ID for auditing.

A key design decision is the \textbf{agent metadata injection}: write tools automatically prefix all posts with structured tags identifying the agent type, agent ID, and execution status. This observability layer enables distinguishing agent-generated posts from human posts, prevents agents from processing their own replies in loops, and supports~filtering.

\subsection{Robot Layer: VLM Controllers and Hardware}
\label{sec:robot_layer}

The Robot Layer encompasses both VLM-driven controllers and the physical robot hardware. A driver module initializes the Unitree SDK over DDS, instantiates the appropriate VLM controller, and routes commands to the correct~robot.

Each VLM controller exposes \textbf{robot primitives} as callable tools. Given a natural language command, the controller runs an \textbf{iterative tool-calling loop}: the VLM receives the command alongside tool definitions, selects and invokes primitives, observes results, and repeats until the command is fulfilled. This lets the VLM decompose complex commands (e.g., ``walk forward then turn around'') into sequences of atomic actions without manual~scripting.

We implement two concrete controllers:

\textbf{Go2 Controller + Unitree Go2~\cite{unitreego2}.} The Go2 is a compact quadruped with 12 degrees of freedom (DOF), equipped with front-facing cameras. The controller exposes four action primitives---\texttt{act\_move} (velocity-controlled locomotion with direction, distance, and speed parameters), \texttt{act\_hello}, \texttt{act\_heart}, and \texttt{act\_backflip}---and two perception primitives for camera capture and cloud image~upload.

\textbf{G1 Controller + Unitree G1.} The G1 is a humanoid robot with 23~DOF, featuring both locomotion and arm manipulation. It exposes \texttt{act\_move} and \texttt{act\_hello} alongside the same perception~primitives.

Both robots communicate via DDS over Ethernet or WiFi through the \texttt{unitree\_sdk2py}~interface.

\section{Agent Design}

\subsection{LLM-Powered Decision Loop}

The core agent follows a perception-reasoning-action loop inspired by the ReAct paradigm~\cite{react}:

\begin{enumerate}[nosep]
    \item \textbf{Perceive}: Poll the forum for new posts via \texttt{list\_posts}, retrieving topic metadata and post content including poster information and timestamps.
    \item \textbf{Reason}: Use the LLM to understand post content, determine if it contains an actionable command directed at this agent, and extract the command if present.
    \item \textbf{Act}: Dispatch the extracted command to the appropriate robot executor, collect the execution result, generate a human-readable summary via the LLM, and post it back to the forum.
\end{enumerate}

Algorithm~\ref{alg:agent} presents the complete agent loop. The agent maintains a set of processed topic IDs to avoid duplicate processing, uses configurable \texttt{@mention} patterns for command targeting, and supports multiple operational modes.

\begin{algorithm}[t]
\caption{AgentRob Main Agent Loop}
\label{alg:agent}
\begin{algorithmic}[1]
\STATE \textbf{Input:} board\_id, poll\_interval, mention\_pattern
\STATE processed $\leftarrow \emptyset$
\STATE Initialize LLM provider, RobotExecutor
\WHILE{running}
    \STATE topics $\leftarrow$ \texttt{list\_posts}(board\_id)
    \FOR{each topic $t$ in topics}
        \IF{$t$.tid $\notin$ processed \AND mentions\_me($t$)}
            \STATE processed.add($t$.tid)
            \STATE cmd $\leftarrow$ LLM.extract\_command($t$.content)
            \IF{cmd $\neq$ null}
                \STATE result $\leftarrow$ Executor.execute(cmd)
                \STATE summary $\leftarrow$ LLM.summarize(cmd, result)
                \STATE \texttt{reply\_to\_topic}($t$.tid, summary)
            \ENDIF
        \ENDIF
    \ENDFOR
    \STATE sleep(poll\_interval)
\ENDWHILE
\end{algorithmic}
\end{algorithm}

\subsection{LLM Provider Abstraction}

All agents access the LLM through a unified \texttt{chat(system\_prompt, user\_message)} interface, regardless of the underlying provider. This abstraction decouples agent logic from any specific LLM service: switching between providers (e.g., Volcengine Doubao, OpenAI, local models) requires only changing configuration, not code. The interface also handles graceful degradation---when the LLM is unavailable, agents fall back to rule-based extraction (Section~\ref{sec:extraction}).

\subsection{Command Extraction}
\label{sec:extraction}

Command extraction transforms unstructured forum posts into actionable robot commands. The primary strategy is \textbf{LLM-based extraction}: a specialized system prompt (Listing~\ref{lst:extraction}) instructs the LLM to identify and extract commands directed at a specific robot agent.

\begin{lstlisting}[
  caption={Command extraction system prompt (Go2). The G1 variant replaces \texttt{@quadruped} with \texttt{@humanoid}.},
  label={lst:extraction},
  language={},
  basicstyle=\ttfamily\small,
  breaklines=true,
  numbers=none,
  xleftmargin=1em,
  xrightmargin=1em
]
You are a command extraction expert.
Extract commands issued to @quadruped
(quadruped robot dog) from forum posts.

Input:  post content in markdown format
Output: extracted commands, concise and
        accurate (bulleted; numbered if
        sequential)

If the post contains no commands for
@quadruped, return an empty string.
\end{lstlisting}

The LLM returns only the extracted command; invalid or empty responses are filtered out. As a fallback, a \textbf{rule-based extractor} captures text following the \texttt{@mention} tag via pattern matching, providing robustness when the LLM is~unavailable.

\subsection{Execution Result Summarization}

After robot execution, the raw result (containing success status, output, and error messages) is transformed into a human-readable forum reply via the LLM. The summarization prompt instructs the model to produce a concise report covering the executed command, key steps, outcome, and any errors. When the LLM is unavailable, the agent falls back to returning the raw result~text.

\subsection{Operational Modes}

Agents support three operational modes: (1)~\textbf{polling mode} (default), where the agent continuously scans the forum at configurable intervals (default: 30\,s); (2)~\textbf{HTTP service mode}, where the agent exposes a REST API for on-demand triggering and external orchestration; and (3)~\textbf{single-run mode}, where the agent performs one scan and exits, useful for debugging and CI/CD~integration.

\subsection{Multi-Agent Architecture}

AgentRob supports multiple concurrent agents with distinct identities and physical embodiments coexisting in the same forum. Each agent is bound to a specific robot and listens for its own \texttt{@mention} trigger (e.g., \texttt{@quadruped} for Go2, \texttt{@humanoid} for G1). When a post mentions multiple robots, each agent independently extracts and executes only the commands directed at~it.

Agents avoid duplicate processing by maintaining a set of already-handled topic IDs. To prevent reply loops---where one agent's post triggers another---all agent-generated posts carry metadata tags that agents recognize and~skip. A rule-based agent variant is provided for pipeline testing without LLM~dependency.

\section{Implementation}

\subsection{MCP Server}

The MCP Server is implemented in TypeScript using the official MCP SDK (\texttt{@model\-context\-protocol/sdk}). Each of the eight tools (Table~\ref{tab:tools}) is defined as a self-contained module consisting of three components: (1)~a \textbf{Zod schema} that validates input parameters at the protocol boundary, (2)~a \textbf{handler} that invokes forum APIs through a shared HTTP client, and (3)~a \textbf{uniform response envelope} wrapping every result in a structured JSON object with success/error status, tool name, payload, and a unique trace~ID.

Write tools automatically inject agent metadata tags into post content before submission, ensuring all agent-generated posts are machine-identifiable. The server communicates with agents via JSON-RPC~2.0 over~stdio.

\subsection{Forum Client}

The forum client abstracts all platform-specific HTTP communication behind a uniform interface. Its design addresses three key challenges of programmatic forum access:

\textbf{Session management.} The client maintains authentication state through cookie jar management and automatic CSRF token acquisition. Write operations require valid CSRF tokens, which are fetched and cached with transparent refresh on~expiry.

\textbf{Runtime identity switching.} A single client instance can switch between forum accounts at runtime by clearing session state and re-authenticating. This enables multi-persona scenarios where one agent process operates under different identities.

\textbf{Registration abstraction.} Account creation supports both privileged (admin API) and unprivileged (public registration) flows, allowing agents to self-provision identities without manual~setup.

\subsection{MCP Client Library}

On the Python agent side, a shared MCP client library spawns the TypeScript MCP Server as a subprocess and manages the full protocol lifecycle---initialization handshake, capability negotiation, and bidirectional message correlation. The client provides typed wrappers for all eight tools, abstracting the JSON-RPC framing into simple async function~calls.

\subsection{Robot Integration}

The \texttt{robot\_command\_driver} module bridges the Agent Layer and Robot Layer. Given a natural language command, it initializes the Unitree DDS communication layer on the specified network interface, instantiates the appropriate VLM controller, and enters the iterative tool-calling loop described in Section~\ref{sec:robot_layer}.

The Go2 controller exposes four action primitives (\texttt{act\_move}, \texttt{act\_hello}, \texttt{act\_heart}, \texttt{act\_backflip}) and two perception primitives (\texttt{get\_front\_image} for camera capture and \texttt{img\_to\_volc} for cloud upload). The G1 controller exposes a subset (\texttt{act\_move}, \texttt{act\_hello}) alongside the same perception~tools.

\section{Discussion}

\subsection{Safety and Ethics}

Opening robot control through public forums raises important safety and ethical considerations that must be addressed:

\textbf{Permission management.} Forum role systems (admin, moderator, registered account, guest) can be mapped to robot operation privilege levels. For example, only accounts with ``robot operator'' role can post in the command board, while all accounts can view execution reports. Our current prototype implements basic agent-side filtering (the agent only processes posts mentioning its specific trigger), but production deployments should integrate with the forum's built-in permission system.

\textbf{Dangerous command detection.} Commands that could cause physical harm (``run into the wall'', ``jump off the table'') should be detected and rejected before execution. This can be implemented as an additional LLM-based safety filter between command extraction and execution, similar to content moderation systems.

\textbf{Rate limiting.} To prevent abuse, the system should enforce per-account rate limits on command submissions, execution frequency caps, and cooldown periods between consecutive robot operations.

\textbf{Identity disclosure.} All agents identify themselves through mandatory metadata tags (\texttt{[agent\_id=...]}) prepended to posts. This ensures that forum participants can distinguish between agent types---commanding agents, robot agents, and monitoring agents---and enables filtering and~analytics.

\textbf{Physical safety perimeter.} Robots should operate within defined physical boundaries, with hardware-level emergency stops that override software commands.

\subsection{Scalability and Extensibility}

AgentRob's modular architecture supports extension along multiple dimensions:

\textbf{Additional robot types.} The executor interface can be implemented for drones, robotic arms, mobile platforms, and any other actuated system. Each executor translates natural language commands into platform-specific control sequences.

\textbf{Alternative forum platforms.} The MCP tool layer abstracts all forum-specific logic. Replacing the current forum backend with Discourse, Reddit, or a custom platform requires only reimplementing the tool handlers while keeping the agent logic unchanged.

\textbf{Enhanced MCP tools.} The current eight tools cover basic forum CRUD operations. Future tools could include: \texttt{upload\_media} for sharing robot camera images, \texttt{search\_posts} for semantic search across forum history, \texttt{manage\_permissions} for dynamic access control, and \texttt{subscribe\_topic} for WebSocket-based real-time notifications (eliminating polling latency).

\textbf{Multi-robot coordination.} Multiple robots can be coordinated through forum threads, where one agent posts a sub-task and another robot's agent picks it up. This enables emergent multi-robot collaboration without centralized planning.

\subsection{Limitations}

We acknowledge several limitations of the current system:

\textbf{Polling latency.} The forum polling mechanism introduces inherent latency (up to one poll interval, default 30 seconds) between command posting and detection. This makes AgentRob unsuitable for real-time robot control tasks requiring sub-second response times. Future work could leverage WebSocket-based push notifications for reduced latency.

\textbf{Network dependency.} Both the agent (for forum access) and the robot (for command reception) require continuous network connectivity. Network interruptions disrupt the pipeline.

\textbf{LLM hallucination risk.} LLM-based command extraction may misinterpret post content, extracting incorrect commands or fabricating commands that were not present. The rule-based fallback mitigates this but at reduced extraction quality.

\textbf{Single-threaded execution.} Each agent currently processes commands sequentially. Concurrent command execution would require careful management of robot state and physical safety constraints.

\textbf{Limited feedback modalities.} Current implementations return only text-based execution reports. Richer feedback---images, video, sensor data visualizations---would significantly enhance interaction quality but requires additional MCP tools and forum media attachment support.

\section{Conclusion}

We have presented AgentRob, a novel framework that bridges online community forums, LLM-powered autonomous agents, and physical robots through the Model Context Protocol. By repurposing forums---one of the oldest and most ubiquitous forms of online communication---as an asynchronous agent-robot interaction channel, AgentRob enables community-scale multi-agent robot orchestration.

Our key technical contributions include: (1) the forum-mediated interaction paradigm, which provides asynchronous, multi-agent, persistent, and observable agent-robot communication; (2) the MCP-based tool framework with eight standardized operations, enabling any MCP-compatible agent to participate in forum communities; (3) the pluggable robot executor architecture with concrete implementations for the Unitree Go2 quadruped; and (4) the multi-agent coordination mechanism allowing agents with diverse physical embodiments to coexist in the same forum.

While current limitations in latency and feedback modalities constrain real-time applications, the framework opens a rich design space for future exploration.

Future directions include: (1) \textbf{multi-modal interaction}, enabling robots to share images, videos, and sensor data on forums; (2) \textbf{inter-robot collaboration}, where robots communicate through forum threads to coordinate complex tasks; (3) \textbf{community-driven robot learning}, where forum discussions provide training signal for robot skill improvement; and (4) \textbf{decentralized robot networks}, built on federated forum platforms where communities govern their own robot fleets.

\bibliographystyle{ACM-Reference-Format}
\bibliography{references}

\end{document}